\pdfoutput=1

\documentclass[11pt]{article}

\usepackage{acl}

\usepackage{times}
\usepackage{latexsym}

\usepackage[T1, T2A]{fontenc}

\usepackage[utf8]{inputenc}

\usepackage{microtype}

\usepackage{enumitem}
\setlist[itemize]{noitemsep, nolistsep}
\setlist[enumerate]{noitemsep, nolistsep}
\usepackage[russian,british]{babel}

\usepackage{substitutefont}
\substitutefont{T2A}{\familydefault}{Tempora-TLF}

\usepackage{algorithm} 
\usepackage{algpseudocode} 

\usepackage{listings}
\usepackage{xcolor}

\definecolor{codegreen}{rgb}{0,0.6,0}
\definecolor{codegray}{rgb}{0.5,0.5,0.5}
\definecolor{codepurple}{rgb}{0.58,0,0.82}
\definecolor{backcolour}{rgb}{0.95,0.95,0.92}

\lstdefinestyle{mystyle}{
    commentstyle=\color{codegreen},
    keywordstyle=\color{magenta},
    numberstyle=\tiny\color{codegray},
    stringstyle=\color{codepurple},
    basicstyle=\ttfamily\footnotesize,
    breakatwhitespace=false,         
    breaklines=true,                 
    captionpos=b,                    
    keepspaces=true,                 
    numbers=left,                    
    numbersep=5pt,                  
    showspaces=false,                
    showstringspaces=false,
    showtabs=false,                  
    tabsize=2
}

\title{The first neural machine translation system for the Erzya language}

\author{David Dale\thanks{\hspace{0.05cm} The research was conducted between the author's employments at Skolkovo Institute of Science and Technology (Skoltech) and at Meta AI.} \\
 \texttt{dale.david@yandex.ru}\\}

\begin{document}
\maketitle
\begin{abstract}
We present the first neural machine translation system for translation between the endangered Erzya language and Russian and the dataset collected by us to train and evaluate it. The BLEU scores are 17 and 19 for translation to Erzya and Russian respectively, and more than half of the translations are rated as acceptable by native speakers. 
We also adapt our model to translate between Erzya and 10 other languages, 
but without additional parallel data, the quality on these directions remains low. 
We release the translation models along with the collected text corpus, a new language identification model, and a multilingual sentence encoder adapted for the Erzya language.
\end{abstract}

\section{Introduction}
Out of the 7 thousand languages spoken around the world, only a minor fraction is covered by machine translation tools. For example, Google Translate\footnote{\url{https://translate.google.com}} supports only 133 languages, and a recent model by \citet{nllb2022} supports 202 languages. Most other languages are often considered ``low-resource'', although some of them have millions of native speakers. In the context of machine translation, the resources that are low are, primarily, parallel and monolingual text corpora. In this work, we create a machine translation system for the previously uncovered Erzya language with only publicly available resources, a very small budget, and limited human efforts. We hope that it will inspire researchers and language activists to enlarge the coverage of existing NLP resources, and in particular, translation systems. 

Our language of choice is Erzya (\texttt{myv}), which is spoken primarily in the Republic of Mordovia, located in the center of the European part of the Russian Federation. The language, along with its close relative Moksha (\texttt{mdf}), belongs to the Mordvinic branch of the Uralic language family. These two languages, although not mutually intelligible \cite{janurik2017erzya}, are often referred to under the common name ``Mordovian''. Erzya has had a written tradition since the beginning of the 19th century \cite{rueter2013erzya}. Its most widely used alphabet is Cyrillic, although there is a Latin-based alternative alphabet\footnote{\url{http://valks.erzja.info} (currently blocked in Russia)}. Erzya has supposedly 300 thousand speakers\footnote{In the 2010 census, 430 thousand people reported speaking Erzya or Moksha, but their proportions are unclear.}, and it is one of the three official languages in Mordovia. According to the UNESCO classification, the Erzya language has a status of ``definitely endangered'' \cite{unesco_atlas}. Some researchers \cite{janurik2017erzya} put it between the levels 6b (``threatened'') and 7 (``shifting'') on the EGIDS scale \cite{lewis2010assessing}, as it is widely used and transmitted between generations in rural communities but is being gradually displaced by Russian in urban areas.  More details about the use of Erzya are given by \citet{rueter2013erzya}, who is also a major current contributor to Erzya NLP resources.

As far as we know, prior to this work, no neural machine translation (NMT) systems for Erzya have been published. To fill this gap, we create and publicly release\footnote{The source code and links to other resources are provided at \url{https://github.com/slone-nlp/myv-nmt}. } the following deliverables:
\begin{itemize}
    \item A language identification model with enhanced recall for Erzya and Moksha languages;
    \item A sentence encoder for Erzya compatible with LaBSE \cite{feng-etal-2022-language};
    \item A small parallel Erzya-Russian corpus and a larger monolingual Erzya corpus;
    \item Two neural models for translation between Erzya and 11 other languages.
\end{itemize}

For translation between Russian and Erzya, we validate our models both by automatic metrics and with judgments  of native speakers. More than half of the translations are rated as acceptable.

\section{Related Work}
Low-resource NLP and, in particular, machine translation, have attracted a lot of attention. Among the recent ambitious projects are \citet{bapna2022building} and \citet{nllb2022} that aim at creating NMT systems for hundreds of languages and rely heavily on collection of large online corpora and transfer learning. Other works, such as \citet{hamalainen2019template}, focus on efficient use of existing vocabularies and morphosyntactic tools to train machine translation systems for very low-resourced languages.

As far as we know, there are no published large parallel corpora or NMT systems for Erzya. \citet{rueter-tyers-2018-towards} develop an Erzya treebank with a few hundred translations to English and Finnish. \citet{arkhangelsky2019} present an Erzya web corpus\footnote{\url{http://erzya.web-corpora.net/}} along with the way it was collected, but the corpus is available only via the web interface. For other published corpora, the situation is similar. There exists a half-finished rule-based machine translation system between Erzya and Finnish\footnote{\url{https://github.com/apertium/apertium-myv-fin}}, and a grammar parser for Erzya\footnote{\url{https://github.com/timarkh/uniparser-grammar-erzya}}. The software package UralicNLP \cite{uralicnlp_2019} supports Erzya among other languages.

There have been a few attempts to transfer machine learning-based NLP resources to Erzya from high-resource languages. \citet{alnajjar2021word} adapt Finnish, English, and Russian word embeddings to Erzya. 
\citet{muller-etal-2021-unseen}, \citet{acs-etal-2021-evaluating} and \citet{wang-etal-2022-expanding} evaluate the performance of multilingual BERT-like models on natural language understanding tasks for new languages, including Erzya. 

None of the works known to us train machine learning-based models that are capable of generating Erzya language.

\section{Methodology and Experiments}

\subsection{Data Collection}
As there are no large open-access corpora for Erzya, we compile Erzya and Erzya-Russian data from various sources:
\begin{itemize}
    \item  12K parallel sentences from the Bible\footnote{\url{http://finugorbib.com}};
    \item 3K parallel Wikimedia sentences from OPUS \cite{tiedemann-2012-parallel};
    \item 42K parallel words or short phrases collected from various online dictionaries;
    \item the Erzya Wikipedia and the corresponding articles from the Russian Wikipedia;
    \item 18 books, including 3 books with Erzya-Russian bitexts\footnote{\url{http://lib.e-mordovia.ru}};
    \item Soviet-time books and periodicals\footnote{\url{https://fennougrica.kansalliskirjasto.fi}};
    \item The Erzya part of Wikisource\footnote{\url{https://wikisource.org/wiki/Main\_Page/?oldid=895127}};
    \item Short texts by modern Erzya authors\footnote{\url{https://rus4all.ru/myv/}};
    \item News articles from the Erzya Pravda website\footnote{\url{http://erziapr.ru}};
    \item Texts found in LiveJournal\footnote{\url{https://www.livejournal.com}} by searching with the 100 most frequent Erzya words.
\end{itemize}
A more detailed account of the data sources is given in Appendix \ref{app:sources}.

After filtering these texts with the language identification model (Section \ref{sec:langid}), we gathered 330K unique Erzya sentences. A bilingual part of the texts was used for mining additional parallel sentences in Section \ref{sec:mining}.

\subsection{Language Identification}
\label{sec:langid}
To make sure that the extra collected data is in the Erzya language, we train a FastText \cite{joulin2016bag} language classifier for the 323 languages present in Wikipedia. The 267 thousand training texts were sampled from Wikipedia with probabilities proportional to $n_{lang}^{1/5}$, where $n_{lang}$ is the size of Wikipedia in that language\footnote{We adopted the idea of temperature sampling
with T=5 from \citet{tran-etal-2021-facebook} and several other works.}. To increase the recall for Erzya and Moksha languages, we augment this training dataset with Erzya and Moksha Bible texts. The resulting model 
has 89\% accuracy and 86\% macro F1 score on the Wikipedia test set (sampled with the same temperature). For Erzya, it has 97\% precision and 82\% recall. Hyperparameters for all trained models are listed in Appendix \ref{app:hyperparameters}.

\subsection{Erzya Sentence Encoder}
\label{sec:encoder}
To compute sentence embeddings, we use an encoder based on LaBSE \cite{feng-etal-2022-language}, with an extended vocabulary. First, we use the BPE algorithm \cite{sennrich-etal-2016-neural} over a monolingual Erzya corpus to add 19K extra merged tokens to the vocabulary. Then, we fine-tune the model on the limited initial parallel data (the Bible, OPUS, and dictionaries): we update only the token embeddings matrix, using the contrastive loss from \citet{feng-etal-2022-language} over computed sentence embeddings. Finally, after collecting more parallel sentences, we fine-tune the full model on a mixture of tasks: contrastive loss over sentence embeddings, standard masked language modeling loss, and sentence pair classification to distinguish correct translations from random pairs.

\subsection{Mining Parallel Sentences}
\label{sec:mining}
When mining parallel sentences, we strive for high precision. To compensate for the questionable quality of our sentence encoder, we apply the following procedure\footnote{For more details on the mining procedure, please read the source code in the repository that we release.}. 

\begin{itemize}
    \item We perform only local mining, i.e. we compare sentences only across paired documents (for Wikipedia and translated books), or within one document (for the web sources).
    \item To evaluate similarity of two sentences, we multiply the cosine similarity between their LaBSE embeddings by the ratio of the length of the shortest sentence to that of the longest one. 
    \item We further penalize the similarities by partially subtracting from them the average similarities of each sentence to its closest neighbors, similarly to using distance margin from \citet{artetxe-schwenk-2019-margin}.
    \item Given two documents in Russian and Erzya, we use dynamic programming to select a sequence of sentence pairs that have the maximal sum of pairwise similarity scores and go in the same order in both documents.
    \item We accept only the sentence pairs with a score above a threshold, which was manually tuned for each source of texts.
\end{itemize}



In total, this approach yielded 21K more unique parallel sentence pairs. The manual inspection found that more than 90\% of them were matched correctly.

\subsection{Training Machine Translation Models}
To benefit from transfer learning, we base our model on the mBART50 model \cite{tang2020multilingual} pretrained on multiple languages, including two Uralic ones (Finnish and Estonian). We extend its BPE vocabulary with 19K new Erzya tokens, using the same method as in Section \ref{sec:encoder}, and add a new \texttt{myv\_XX} language code to it. Embeddings for the new tokens are initialized as the averages of the embeddings of the Russian tokens aligned with them in the parallel corpus\footnote{We compute alignments with a naive formula: the alignment weight between tokens $t_i$ and $t_j$ is estimated as $\frac{n_{ij}^2}{n_jn_j}$, where $n_i$ and $n_j$ are their respective frequencies in the parallel corpus, and $n_{ij}$ is the number of sentence pairs with $t_i$ in one sentence and $t_j$ in another.}, inspired by \citet{xu-hong-2022-sub}.

We make two copies of this model and train them to translate in the \texttt{myv-ru} and \texttt{ru-myv} directions, respectively. The \texttt{myv-ru} model is trained on the joint parallel corpus of sentences and words. The \texttt{ru-myv} model is trained on the union of this corpus and the back-translated corpus generated by the \texttt{myv-ru} model from the monolingual \texttt{myv} data. 

After training the models on these two languages, we adapt them to 10 more languages: \texttt{ar}, \texttt{de}, \texttt{en}, \texttt{es}, \texttt{fi}, \texttt{fr}, \texttt{hi}, \texttt{tr}, \texttt{uk}, and \texttt{zh}, resulting in the \texttt{myv-mul} and \texttt{mul-myv} models (below, by \texttt{mul} we denote any of these 10 languages). We fine-tune the two models jointly, using a version of online-back translation and self-training. Specifically, we generate the training pairs in four alternating steps:
\begin{enumerate}
    \item Sample a \texttt{ru-mul} sentence pair from the CCMatrix \cite{schwenk-etal-2021-ccmatrix} dataset, translate from \texttt{ru} to \texttt{myv} with the \texttt{mul-myv} model;
    \item Sample a \texttt{ru-mul} pair from the CCMatrix, translate from \texttt{mul} to \texttt{myv} with the \texttt{mul-myv} model;
    \item Sample a \texttt{ru-myv} pair from our parallel corpus, translate from \texttt{myv} to \texttt{mul} with the \texttt{myv-mul} model;
    \item Sample a \texttt{myv} text from the monolingual \texttt{myv} corpus, translate from \texttt{mul} to \texttt{myv} and \texttt{ru} with the \texttt{myv-mul} model.
\end{enumerate}
At each step, we update both models on the \texttt{myv-mul} and \texttt{myv-ru} pairs in both directions. For the self-training updates, we multiply the loss by the coefficient $\lambda_{ST}=0.05$ to decrease the impact of self-training relatively to back-translation (the choice of the coefficient is suggested by experiments in \citet{he-etal-2022-bridging}).

During the initial experiments, we noticed that, when translating from Russian to Erzya, the model often just copied Russian phrases with only word endings sometimes changed. Sometimes this is acceptable because Erzya has multiple Russian loanwords, but often there exist native words that are preferable. To alleviate this problem, in step 1 we generate 5 alternative \texttt{ru-myv} translations using diverse beam search \cite{vijayakumar2016diverse}, and choose the one with the largest proportion of words recognized as \texttt{myv} by our language identification model. This problem was also the reason why we chose to train two different models from translation to Erzya and from Erzya: this way, the decoder and encoder of a model never work with the same language.

\section{Evaluation}
\subsection{Data}
For model evaluation, we prepare a held-out corpus of 3000 aligned Erzya-Russian sentences from 6 diverse sources: the Bible, Erzya folk tales \cite{sheyanova2017}, the Soviet 1938 constitution, descriptions of folk children's games \cite{bryzhinsky2009}, modern Erzya fiction and poetry, and Wikipedia. To evaluate English and Finnish translation, we use translations from the Erzya universal-dependency treebank \cite{rueter-tyers-2018-towards}: 441 sentence pairs for \texttt{en}, and 309 for \texttt{fi}. We split all these sets into development and test parts, and report the results on the test set.

\subsection{Automatic Metrics}
For all evaluated directions (between \texttt{myv} and \texttt{ru, en, fi}) we calculate BLEU \cite{papineni-etal-2002-bleu} and ChrF++ \cite{popovic-2017-chrf}. Both these metrics estimate the proportion of common parts in the translation and the reference, but BLEU is calculated as precision over word n-grams, whereas ChrF++ aggregates precision and recall of word and character n-grams (which is more suitable for morphologically rich languages such as Erzya and Russian).
The values of these metrics on the test set are given in Table \ref{tab:bleu}. For translation from and to Russian, the BLEU scores are 17 and 19 points, respectively. For English and Finnish, however, BLEU is well below 10. We hypothesize that the low quality may be attributed to the domain mismatch between the Erzya-origin and English- or Finnish-origin training corpora, but without detailed test sets we cannot verify this.

\begin{table}
\small
\centering
\begin{tabular}{l|rrr}
\hline
\textbf{Direction} & \textbf{BLEU} & \textbf{ChrF++} \\
\hline
ru-myv & 17.71 & 41.16 \\
myv-ru & 19.68 & 38.63 \\
en-myv & 2.77 & 28.03 \\
myv-en & 5.44 & 25.99 \\ 
fi-myv & 4.79 & 27.42 \\
myv-fi & 3.02 & 22.34 \\
\hline
\end{tabular}
\caption{\label{tab:bleu} Reference-based scores on the test sets.}
\end{table}

\begin{table}
\small
\centering
\begin{tabular}{l|rr|rr}
\hline
 & \multicolumn{2}{c|}{ru-myv} & \multicolumn{2}{c}{myv-ru} \\
 \hline
\textbf{Source} & \textbf{BLEU} & \textbf{ChrF++} & \textbf{BLEU} & \textbf{ChrF++} \\
\hline
bible                & 10.00 & 36.92 & 10.71 & 33.55 \\
tales                &  7.00 & 33.90 &  7.30 & 28.42 \\
constitution         & 27.82 & 62.96 & 33.31 & 60.60 \\
games                & 10.33 & 31.19 &  9.85 & 26.57 \\
fiction              &  8.68 & 30.59 &  5.95 & 26.60 \\
wiki                 & 28.39 & 48.56 & 32.24 & 47.55 \\
\hline
\end{tabular}
\caption{\label{tab:bleu_ru} Scores by section on the \texttt{myv-ru} test set.}
\end{table}

For the Russian test set, the performance varies greatly depending on the domain (Table \ref{tab:bleu_ru}). The constitution has the highest scores because its Erzya text is saturated with Russian loanwords and is easy to generate and understand. For Wikipedia, the scores are also high, probably because its Erzya articles are often translated from Russian in a rather literal way. The other domains have a more artistic style, and the translations are on average much less literal.

Some examples of the translations and references are given in Table \ref{tab:examples}.

\begin{table*}[ht!]
\small
\centering
\begin{tabular}{|l|p{0.7\linewidth}|}
\hline
Type                 & Text                                                                                                                                   \\
\hline
source (ru)          & И   вот что рассказывают наши русские старики: «Когда здесь не было этого села, в   этом овраге были разбойники; у них были землянки.  \\
source (myv)         & Вана   мезе ёвтнить миненек рузонь атятне: зярдо велесь тесэ арасель, се латксонть   эрясть розбойникть, эрясть землянкасо.            \\
translation (ru-myv) & Ды   вана мезе ёвтнить минек эрзянь атятне: «Зярдояк те велесэнть арасель, тосо   оврагасо ульнесть розбойникть; сынст ульнесть землянкаст. \\
translation (myv-ru) & Вот что нам говорят русские   старцы: когда деревня здесь не была, то там жилибойники, жили на земле.                                  \\
\hline
source (myv)         & Кода   авазо, анокстась лапужа кирькст, истя жо педявтнинзе, валаськавтнинзе   педявтома таркатнень начко кедьсэ.                      \\
source (en)          & Like his mother, he prepared   flat rings, and stuck them onto the patty in the same way, and smoothed out   the seams with his wet hands.  \\
source (fi)          & Samalla tavalla kuin äitinsä   Ketšai valmisti litteitä rinkuloita liitti ne samalla tavalla, ja siloitti   liitoksen märällä kädellä. \\
translation (myv-en) & Like his mother, he prepared   flat circles, and also filled the canvas with a needle.                                                 \\
translation (myv-fi) & Kuten äiti, valmistelee   tasa-alaiset kentät, myös venytetään, lyödään venyttäjän käsillä.                                            \\
translation (en-myv) & Кода аванзо, сон анокстыль   валаня суркст, теке ладсо педявтызе сынст пацьказонзо ды вадяшась кедень   летькенть марто.               \\
translation (fi-myv) & Истя жо, кода авазо Кетшай   анокстыль лаҥгсо кевпанть, сон солодиль сынст теке ладсо ды солодиль эйсэст   кедьлапушкасо.    \\
\hline
\end{tabular}
\caption{\label{tab:examples} A few examples of translations and references.}
\end{table*}

\subsection{Manual Evaluation}
\label{sec:manual}
We recruit three native speaker volunteers to evaluate some translations manually. The evaluation protocol is similar to XSTS \cite{nllb2022}, but evaluates fluency in addition to semantic similarity. The scores are between 1 (a useless translation) and 5 (a perfect translation), with 3 points standing for an acceptable translation without serious errors. Criteria for each score are given in Appendix \ref{app:guidelines}.

Each of the 3 annotators rated a few randomly sampled translations from the dev split of each source: 12 pairs in the \texttt{ru-myv} and 17 pairs in the \texttt{myv-ru} directions, which amounts to 87 sentence pairs annotations in total. The average length of the labelled texts was 97 characters, or 14 words.

It turned out that, despite the specified annotation criteria, the annotators were calibrated very differently: their average ratings were 2.9, 3.5, and 4.1. We chose a pessimistic aggregation strategy: for each of the 29 evaluated sentence pairs, we took the worst of the scores by our 3 volunteers.

For translation to Erzya, the average pessimistic score was 2.75, and 58\% translations were rated as acceptable (i.e. all the 3 reviewers rated them with at least 3 points). For translation to Russian, the average score was 2.71, with 53\% acceptable translations.

An additional comment from the annotators was that some of the source Erzya texts were inadequate. In particular, some \texttt{games} sentences contained grammatical errors\footnote{We are not certain whether these errors are due to the low quality of the source text, or to the natural variations within the Erzya language.}, and most \texttt{constitution} sentences contained Russian words with Erzya endings instead of their Erzya equivalents. This suggests that one of the next steps in improving our NMT system might be to filter the training and evaluation data for better language quality.

\section{Conclusions and Future Work}
In this paper, we present the first NMT system for the endangered Erzya language, capable of translating between it and 11 diverse languages, primarily Russian. During its development, we have collected about 30K parallel Russian-Erzya sentences and 300K monolingual Erzya sentences, and trained a language identification model and a BERT-based sentence encoder that support Erzya. All the resources are publicly released. These efforts have occupied about two man-weeks of working time and almost no expenses\footnote{All the expenses incurred totalled \$9.99 for the paid subscription to the Google Colab system (\url{https://colab.research.google.com/signup}).}. We hope that these results will inspire the NLP community to develop resources for other endangered languages.

The quality of our system may be improved by collecting more texts in Erzya and filtering them better than we did. Another promising direction is a more efficient usage of the vocabularies and parsers that are already available for the language, e.g. for generating synthetic training data. Finally, we hope to attract more native speakers for creating larger and cleaner train and test datasets.

One open research question is that of transfer between languages: whether Erzya translation benefits from knowledge of, for example, Hungarian or Estonian, and whether knowledge of Erzya can bring improvements to other languages, such as Moksha. In further studies, we hope to shed some light on this direction as well.

\section{Acknowledgements}
We gratefully acknowledge support from the volunteers who participated in the manual evaluation of translation quality: Semyon Tumaikin, Zinyoronj Santyai, and Evgenia Chugunova. We are also grateful to the reviewers for their suggestions which helped to improve this work.

\bibliography{anthology,custom}
\bibliographystyle{acl_natbib}

\appendix
\onecolumn

\section{Data sources}
\label{app:sources}

\begin{table*}[ht!]
\begin{tabular}{|p{0.7\linewidth}|l|r|}
\hline
Source                          & Type           & Size   \\
\hline
Erzya-Russian dictionaries: \citet{marlamuter}, \citet{mordovians},   \citet{mordvarf},  \citet{ryabov2021},   \citet{schankina2011}, \citet{erushov} & phrase pairs   & 47860 \\
\hline
The \texttt{myv-ru} Wikimedia corpus on   OPUS \cite{tiedemann-2012-parallel}                                             & sentence pairs & 3202  \\
The Bible \cite{finugorbib}                                    & sentence pairs & 12483  \\
\citet{sheyanova2017} (aligned)                                & sentence pairs & 1023   \\
\citet{bryzhinsky2009} (aligned)                               & sentence pairs & 4203   \\
Erzya and Russian Wikipedia   \cite{wikidump} (aligned)        & sentence pairs & 11479  \\
Livejournal \cite{lj} (aligned)                                & sentence pairs & 1799   \\
Modern Erzya fiction and poetry   \cite{rus4all} (aligned)     & sentence pairs & 916    \\
The Soviet 1938 constitution \cite{const}   (aligned)          & sentence pairs & 304    \\
Mordovian tales and riddles   \cite{evsenyev} (aligned)        & sentence pairs & 3776   \\
\hline
Various Erzya fiction books   \cite{emordovia}                 & sentences      & 52870  \\
Various Soviet-time books and periodicals   \cite{fennougrica} & sentences      & 54798  \\
Erzya Wikisource, filtered by language   \cite{wikisource}     & sentences      & 120470 \\
Articles from the Erzya Pravda website   \cite{pravda}         & sentences      & 43772  \\
Livejournal \cite{lj}                                          & sentences      & 36584  \\
Erzya Wikipedia \cite{wikidump}                                & sentences      & 59569  \\
\citet{bryzhinsky2009}                                         & sentences      & 5194    \\
\hline
\end{tabular}
\caption{\label{tab:sources} The sources used to construct the training and evaluation datasets. The ``size'' column denotes the number of sentences or phrases in the source.}

\end{table*}


\section{Models' hyperparameters}
\label{app:hyperparameters}

\subsection{Language identification}
For the language identification model, we use the official FastText implementation\footnote{\url{https://github.com/facebookresearch/fastText}}. We train it with initial learning rate of 0.05 for 100 epochs, using minimum word count of 100, 64-dimensional embeddings and 200K hash buckets for character n-grams with n from 1 to 4. Then we quantize the model with retraining on the same dataset, a cutoff of 50000, and norm pruning.

\subsection{Sentence encoder}
For the sentence encoder model, we use a PyTorch port of LaBSE\footnote{\url{https://huggingface.co/sentence-transformers/LaBSE}}, in which we remove tokens for all languages, except Russian and English, and add Erzya tokens. For vocabulary extension, we set the minimal token count for stopping BPE at 30. 

After extending the vocabulary, we fine-tune the model on the initial parallel sentences and phrases using the LaBSE contrastive loss with margin 0.3 and batch size 4 for 500K steps, updating only the embeddings, and passing the gradient only through the encoded \texttt{myv} sentence. We use the Adafactor optimizer with learning rate of $10^{-5}$ and clipping the gradient norm at 1. Then we update the model for 500K steps with learning rate $2 \times 10^{-6}$, updating all the parameters, and alternating batches with the LaBSE loss, MLM loss, and the loss of classifying the correct and incorrect sentence pairs. Incorrect pairs are generated either by sampling one of the sentences randomly, or by randomly inserting, deleting, or swapping words in one of the sentences in a correct parallel pair.

\subsection{Machine translation models}
Both \texttt{myv-ru} and \texttt{ru-myv} models were initialized from mBART50\footnote{\url{https://huggingface.co/facebook/mbart-large-50-many-to-many-mmt}} with the vocabulary extended with Erzya tokens. They were trained with Adafactor optimizer using batch size of 8 and learning rate of $10^{-6}$ for 4 epochs: on the first epoch, only token embeddings were updated, and on the remaining epochs, all parameters were updated. 

The \texttt{myv-mul} and \texttt{mul-myv} models were initialized from \texttt{myv-ru} and \texttt{ru-myv}, respectively. They were jointly trained for 40K updates with batch size of 1. 

For inference, we used beam size of 5 and repetition penalty of 5.0.

Both the sentence encoder and the translation models were trained using the PyTorch\footnote{\url{https://pytorch.org}} and Transformers\footnote{\url{https://huggingface.co/docs/transformers/}} Python packages.


\section{Quality annotation guidelines}
\label{app:guidelines}

The following annotation criteria (in Russian) were suggested to the annotators in Section \ref{sec:manual}.

\begin{itemize}
    \item 5 points: a perfect translation. The meaning and the style are reproduced completely, the grammar and word choice are correct, the text looks natural. 
    \item 4 points: a good translation. The meaning is reproduced completely or almost completely, the style and the word choice are natural for the target language.   
    \item 3 points: an acceptable translation. The general meaning is reproduced; the mistakes in word choice and grammar do not hinder understanding; most of the text is grammatically correct and in the target language.
    \item 2 points: a bad translation. The text is mainly understandable and mainly in the target language, but there are critical mistakes in meaning, grammar, or word choice. 
    \item 1 point: a useless translation. A large part of the text is in the wrong language, or is incomprehensible, or has little relation to the original text. 
\end{itemize}

\end{document}